# Machine Learning Classification of Cryopathy Syndromes: A Comprehensive Comparative Study

Nataliya Shakhovska, Lviv Polytechnic National University, 79013, Lviv, Ukraine, nataliya.b.shakhovska@lpnu.ua;
Valentyna Chopyak, Danylo Halytskyi Lviv National University, 79010 Lviv, Ukraine; chopyakv@gmail.com;
Ivan Izonin, Lviv Polytechnic National University, 79013, Lviv, Ukraine, ivan.v.izonin@lpnu.ua;
Vira Haievska, Danylo Halytskyi Lviv National University, 79010 Lviv, Ukraine, vira00017@gmail.com.
Corresponding author: Ivan Izonin, ivan.v.izonin@lpnu.ua

***Abstract.***

*Background:* Cryopathy syndromes are difficult to classify because laboratory patterns often overlap across diagnostic categories, while some diagnoses are rare. This makes routine interpretation of cryoglobulin-related tests challenging and increases dependence on expert judgment. The aim of this study was to develop and compare machine learning approaches for automated classification of cryopathy syndromes from laboratory data and to identify a practical strategy for clinical decision support.

*Methods:* We analysed laboratory records from 2,686 patients assigned to 14 diagnostic categories. The dataset included demographic variables, cryoglobulin measurements, precipitation tests, and hemagglutinin and hemolysin titers. Data preprocessing included cleaning, encoding, imputation, normalization, and construction of clinically informed interaction features. We evaluated 12 modelling strategies, including Random Forest, Gradient Boosted Trees, Multi-Layer Perceptron, soft-voting ensembles, class balancing with Synthetic Minority Over-sampling Technique, hierarchical classification, period-aware models, targeted binary classifiers, and probability calibration. Performance was assessed using stratified train-test evaluation and stratified 5-fold cross-validation. The main metrics were macro-averaged F1 score, accuracy, Cohen’s kappa, Top-3 accuracy, and expected calibration error.

*Results:* The overall task proved difficult because of marked class imbalance and clinical overlap between diagnoses. The best multiclass performance was achieved by a soft-voting ensemble of Random Forest and Gradient Boosted Trees. Cross-validation confirmed stable performance for the balanced Random Forest model. Tree-based methods consistently outperformed the neural network model. Feature engineering improved discrimination, and the most informative predictors were derived cryoglobulin-based interaction features. Probability calibration substantially improved reliability for the largest class. Binary classifiers for selected diagnostic pairs performed much better than the full multiclass models, with the strongest results observed for Hepatitis C versus Chronic Hepatitis and for Cryovasculitis versus immunodeficiency-related disease.

*Conclusions:* Machine learning can capture diagnostically relevant structure in cryoglobulin laboratory data, although full multiclass classification remains difficult and the present results should be interpreted as exploratory. A more realistic near-term use of the proposed approach is as a preliminary decision-support prototype for shortlist generation, combining calibrated multiclass screening with targeted binary classification for difficult differential diagnosis pairs.

*Trial registration:* Not applicable.

**Introduction.**

Cryopathies comprise a heterogeneous group of clinical conditions associated with abnormal cold-sensitive proteins in the blood, known as cryoglobulins. These immunoglobulins reversibly precipitate at temperatures below 37°C and may lead to a broad spectrum of manifestations, ranging from cutaneous vasculitis and peripheral neuropathy to severe systemic organ involvement. Their clinical importance extends beyond cryoprecipitation itself, since cryoglobulins also act as immunological markers of underlying disorders, including chronic viral infections, autoimmune diseases, and lymphoproliferative malignancies [1], [2].

The differential diagnosis of cryopathy syndromes remains particularly difficult because laboratory findings often overlap across pathogenetic types. Autoimmune cryopathies may share elevated cryoglobulin levels and positive autohemagglutinin titers, yet differ in the pattern of immunoglobulin involvement and complement activation. Immunodeficiency-related cryopathy, which was the most frequent diagnosis in our cohort, can present with a broadly similar serological profile despite a different underlying mechanism. Allergic and mediator-type cryopathies also show related clinical manifestations, although they arise from different biological processes. Viral-associated conditions likewise partially overlap with autoimmune forms. As a result, standard laboratory panels provide only a limited view of the full immunological state, and a single laboratory profile may be compatible with several diagnostic categories [1], [2].

In current clinical practice, the diagnosis of cryopathies depends largely on the expert judgement of immunologists, who interpret laboratory findings together with clinical symptoms, medical history, and imaging results. Although this approach remains clinically valuable, it is also time-consuming, partly subjective, and difficult to standardize across institutions. The absence of validated algorithmic criteria for cryopathy subtyping therefore creates a strong rationale for the use of machine learning, which may reveal multivariate patterns in laboratory data that are not readily captured by conventional interpretation. This perspective is consistent with recent studies showing that machine learning can support diagnostic reasoning in autoimmune and other clinically complex settings [3], [4].

Cryopathies are especially relevant in populations exposed to prolonged cold stress. In Ukraine, this issue has become even more pressing in the context of the Russo-Ukrainian war, which has increased the number of individuals exposed to severe field conditions and, consequently, the clinical burden of cold-sensitivity disorders. The present dataset, collected at a Ukrainian clinical centre across the pre-COVID, COVID, and wartime periods, reflects this tendency and underscores the need for efficient and objective diagnostic support tools. More broadly, this need aligns with current trends in medical informatics, where transparent machine learning models and clinically oriented decision-support tools are increasingly developed for real-world diagnosis [5], [6].

Although machine learning has shown considerable promise in clinical diagnostics [7], [8], its application to cryopathy classification poses several specific challenges. These include marked class imbalance [7], a limited number of available laboratory features [9], and genuine clinical overlap between diagnostic categories [10]. Against this background, the present study aimed to develop and compare a broad set of classification approaches, including tree-based ensembles, neural networks, and voting-based models; to evaluate the contribution of class balancing, feature engineering, and hyperparameter optimisation; to investigate hierarchical, period-aware, and binary diagnostic strategies; to obtain calibrated probability estimates for clinical risk assessment; and to validate the results by rigorous cross-validation. The relevance of such a design is supported by recent evidence that machine learning methods can improve diagnostic modelling in autoimmune and laboratory-based clinical tasks when carefully integrated with domain knowledge [4], [11].

To the best of our knowledge, we did not identify prior studies that addressed multiclass machine-learning classification of cryopathy syndrome categories from routine cryoglobulin-related laboratory data. Although machine learning has already been applied to autoimmune and rheumatic diseases, including related biomarker-based diagnostic tasks, no published study has specifically addressed the computational classification of cryoglobulinemia-associated conditions. In this sense, the present work fills an important gap and establishes an initial quantitative baseline for future studies in this area [4], [12].

In this paper, we present an application-oriented combination of modelling strategies for a highly imbalanced multiclass clinical task. Rather than relying on a single model, we combine three complementary strategies: calibrated multiclass ensemble screening with Top-3 candidate shortlisting, targeted binary classifiers for selected differential diagnosis pairs, and probability outputs refined by Platt scaling. This design is clinically meaningful because it mirrors the actual diagnostic process, in which broad initial screening is followed by more focused differential assessment under uncertainty. Such an approach is consistent with the broader movement toward clinically applicable and explainable decision-support systems in medical informatics [3], [6].

A further contribution of this study is the empirical analysis of cryopathy diagnostic patterns across three epidemiological periods: pre-COVID, during COVID-19, and during active military conflict. The observation that the mapping between cryoglobulin-related features and diagnosis remains broadly stable across these periods is important, because it suggests a degree of diagnostic consistency despite changes in patient flows, healthcare access, and the wider clinical environment. The wartime subset is particularly valuable in this regard, as it captures an emerging burden of cold-sensitivity disorders among military personnel and civilians exposed to prolonged field conditions.

In addition, we identify and empirically assess a set of domain-informed engineered features, including the interaction between cryoglobulin level and autohemagglutinin titer, the auto-to-iso hemagglutinin ratio, and the logarithmically transformed cryoglobulin level. These features outperformed raw laboratory variables in permutation-based importance analysis, which indicates that clinically informed feature construction can meaningfully strengthen predictive modelling. This observation is in line with recent work showing that biomarker-informed and interpretable machine learning approaches can improve diagnostic performance in complex medical settings [13], [14].

All ML algorithms in this study were implemented from scratch in NumPy, including Synthetic Minority Over-sampling Technique, Random Forest, Gradient Boosted Trees, Multi-Layer Perceptron, soft-voting ensembles, Platt scaling, and stratified cross-validation. This choice ensured a transparent and reproducible methodological framework and reduced dependence on library-specific defaults or hidden heuristics, which can complicate fair comparison between models. In the context of medical informatics, where interpretability, reproducibility, and methodological clarity remain especially important, such transparency is a substantive strength rather than merely a technical detail [5], [15].

## Background and Related Studies.

### *2.1 Cryoglobulinemia: Clinical Background and Diagnostic Challenges.*

Cryoglobulinemia is defined by the presence of immunoglobulins in the blood that reversibly precipitate when the temperature falls below 37°C. According to the classical Brouet classification, three main types are distinguished: Type I, which is associated with monoclonal immunoglobulins, usually IgM or IgG; Type II, which includes monoclonal IgM together with polyclonal IgG and is typically referred to as mixed cryoglobulinemia with rheumatoid factor activity; and Type III, which is characterized by polyclonal IgG and IgM [16].

Recent clinical literature has further clarified the heterogeneity of this condition. In particular, a recent review in The New England Journal of Medicine described cryoglobulinemia as “one name for two diseases,” thereby emphasizing the profound pathophysiological distinction between Type I forms and mixed cryoglobulinemia [17]. A more recent review by Zignego et al. also highlighted that, despite progress in understanding classification, pathophysiology, and treatment, the detection and typing of cryoglobulins remain insufficiently standardized in routine clinical practice [2].

This lack of standardization is not only a laboratory issue. It also directly affects diagnostic consistency and reproducibility. In such conditions, machine learning appears to be a reasonable and timely direction, as it may support a more objective interpretation of complex laboratory patterns.

### *2.2 Machine Learning in Autoimmune and Rheumatic Disease Diagnosis.*

In recent years, machine learning has been increasingly applied to the diagnosis of autoimmune and rheumatic diseases. Kruta et al. proposed an integrated framework that combined multi-omics data with clinical laboratory parameters and reported high predictive performance across several machine learning models, including logistic regression, random forest, neural networks, and support vector machines [4].

Other studies have also confirmed the promise of this direction. A large retrospective study involving more than twelve thousand patients evaluated several modern classifiers, including Random Forest, LightGBM, XGBoost, CatBoost, and TabNet, for calibrated and explainable multiclass diagnosis of rheumatic diseases [18]. In primary care screening for rheumatoid arthritis, CatBoost showed particularly strong performance, indicating that machine learning can be useful even at early stages of clinical decision-making [19]. Similar conclusions were drawn in studies on early-stage systemic autoimmune rheumatic diseases, where multiclass models based on basic laboratory indicators demonstrated encouraging results across several disease categories [20].

At the same time, an important gap remains. Despite the growing number of studies in autoimmune disease classification, we did not identify published work specifically devoted to machine learning-based classification of cryoglobulinemia or cryopathy syndromes. This makes the present problem both scientifically relevant and insufficiently explored.

### *2.3 Machine Learning for Related Cold-Sensitivity Conditions.*

Machine learning has also been applied to clinically related cold-sensitivity disorders, particularly Raynaud's phenomenon. Kaczmarek et al. developed convolutional neural networks for the differential diagnosis of Raynaud's phenomenon from hand thermal images and showed that meaningful discrimination can be achieved even without a cold stimulation procedure [21].

More recently, the ARTIX approach proposed a machine learning-based index for Raynaud's assessment using mobile phone photography. Its validity was supported by the ability to quantify finger response during a standardized cold challenge [22].

These studies are important because they demonstrate that artificial intelligence methods can indeed support the evaluation of cold-related disorders. However, they rely mainly on imaging data. In contrast, laboratory-based classification of cryopathy syndromes remains largely unaddressed.

### *2.4 Tree-Based Ensembles for Tabular Clinical Data.*

For tabular clinical data, tree-based ensemble methods still remain among the most reliable modelling approaches. Random Forest provides robust predictions by combining bootstrap aggregation with random feature selection, whereas Gradient Boosted Trees construct a sequence of models that gradually correct previous errors [23].

This advantage has been repeatedly confirmed in comparative studies. For example, Yang et al. showed that gradient-boosting decision tree models, including XGBoost, LightGBM, and CatBoost, generally outperform deep learning alternatives on a range of medical tabular datasets, while also remaining easier to optimise and less computationally demanding [24]. A broader review of clinical machine learning applications similarly found that Random Forest is one of the most frequently used methods in medical data analysis, alongside logistic regression and support vector machines [25].

Taken together, these findings suggest that tree-based ensembles should be considered a strong baseline, and often a preferred solution, for structured clinical datasets of moderate size.

### *2.5 Deep Learning for Tabular Data.*

Deep learning methods for tabular data have attracted considerable attention in recent years. Architectures such as TabNet and FT-Transformer were proposed to improve feature representation and selection in structured datasets [26]. In some benchmark settings, transformer-based approaches achieved competitive or even superior performance.

However, the overall picture remains nuanced. Although newer neural architectures are promising, the broader empirical consensus is still that tree-based models usually perform better on tabular datasets, especially when the data are limited in size and contain a mixture of numerical and categorical variables [27]. This observation is particularly relevant for medical applications, where datasets are often relatively small, class distributions are imbalanced, and interpretability remains important.

For this reason, deep learning is best viewed here not as an automatic replacement for classical machine learning, but as an alternative that should be evaluated critically against strong ensemble baselines.

### *2.6 Handling Class Imbalance in Medical Data.*

Class imbalance is one of the central challenges in medical machine learning. This is especially true in diagnostic problems, where some disease categories are represented by many observations, while others remain rare. Under such conditions, standard classifiers tend to favour majority classes and may fail to recognise clinically important minority groups.

One of the most widely used strategies for addressing this issue is Synthetic Minority Over-sampling Technique. This method generates additional minority samples through interpolation between neighbouring observations [28]. At the same time, conventional Synthetic Minority Over-sampling Technique also has limitations, since synthetic points may be placed in ambiguous regions of the feature space and may therefore reduce class separability.

To overcome these issues, a number of modified oversampling strategies have been proposed. These include Distance-based Synthetic Minority Over-sampling Technique, Bi-phasic Synthetic Minority Over-sampling Technique, and more recent multiclass methods that explicitly account for class-specific neighbourhood structure [29]. Other studies have shown that, for imbalanced medical datasets, the best results are often achieved not by oversampling alone, but by combining balancing procedures with ensemble learning [30]. Adaptive variants developed specifically for healthcare applications further confirm the importance of tailoring imbalance-handling strategies to the structure of the diagnostic problem [31].

### *2.7 Probability Calibration and Clinical Decision Support.*

In clinical applications, predictive accuracy alone is not sufficient. It is equally important that the predicted probabilities reflect real diagnostic uncertainty. This is why probability calibration has become a central topic in medical machine learning.

A commonly used approach is Platt scaling, which transforms raw model scores into calibrated probabilities through a sigmoid function. Another widely discussed alternative is isotonic regression, which provides a more flexible non-parametric calibration scheme [32]. Comparative studies suggest that the effect of calibration may depend on the model and the data. In some cases, Platt scaling improves reliability, while in others isotonic regression performs better, especially when larger calibration sets are available [33].

For this reason, calibration quality should be assessed explicitly rather than assumed. In current clinical prediction research, Expected Calibration Error is one of the most widely used measures, since it quantifies the discrepancy between predicted confidence and observed accuracy [34]. From the perspective of decision support, this is especially important because clinicians need not only a predicted class, but also a trustworthy estimate of how confident the model is in that prediction.

### *2.8 Hierarchical and Cascaded Classification.*

Hierarchical classification offers another promising direction for complex multiclass medical problems. Its key idea is to decompose a difficult classification task into a sequence of simpler and clinically meaningful decisions.

In medicine, this approach naturally corresponds to the way clinicians often reason. A physician may first identify a broader pathogenetic category, such as autoimmune, viral, or allergic disease, and only then move to a more specific diagnosis within that group. From a machine learning perspective, such decomposition may reduce the effective complexity of the task and improve discrimination when direct multiclass classification is too difficult.

This idea has already been explored in rare disease diagnostics and other settings with limited samples per class [35]. For cryopathy syndromes, hierarchical modelling appears especially relevant because the diagnostic structure itself is layered, and some categories are much easier to distinguish at the broader mechanistic level than at the level of specific final diagnoses.

**Methods.**

***3.1 Dataset.***

The dataset comprised 2,686 patient records collected at a clinical immunology laboratory in Lviv, Ukraine, across three epidemiological periods: pre-COVID, during COVID, and during wartime. The wartime subset is of particular clinical interest because the Russo-Ukrainian war has created a substantial population of military personnel and civilians exposed to prolonged cold stress in field conditions, including trenches, open terrain, and unheated shelters.

Each record included demographic variables, namely age and sex, as well as laboratory measurements relevant to cryopathy assessment. These variables included the spectrophotometric cryoglobulin level, which served as the main quantitative marker of cryoglobulin concentration in serum, binary precipitation and paraprotein tests, and several titer-based indicators reflecting the intensity of cold-reactive immunological responses. The examination period was also retained as a categorical variable and later encoded for period-aware modelling.

The original dataset contained 18 diagnostic categories. Classes represented by fewer than 35 cases were merged into an Other category, which resulted in 14 final classes. The class distribution was highly imbalanced. The most frequent diagnosis, immunodeficiency-related disease, accounted for 722 cases, whereas the rarest retained classes contained fewer than 40 cases. This produced an imbalance ratio exceeding 80:1. The final diagnostic structure covered six broader pathogenetic groups: autoimmune, immunodeficiency-related, allergic, mediator-type, viral-associated, and control. This taxonomy was later used in the hierarchical classification framework.

***3.2 Preprocessing and Feature Engineering.***

The raw clinical data required substantial preprocessing in order to convert heterogeneous medical notation into a numerical representation suitable for machine learning. The preprocessing pipeline included four main stages: data cleaning, variable encoding, feature engineering, and normalization.

In the data cleaning stage, titer values originally recorded as clinical ratios, such as 1:64 or 1:128, were parsed and converted into their numeric denominators. These values reflect the magnitude of the immunological response and are more convenient for computational analysis. When multiple titers were present in a single entry, the maximum value was retained. Qualitative binary test results were encoded as 1 or 0 from their original Ukrainian responses. Missing values in numerical variables were imputed using the median of the corresponding feature. This choice is robust to outliers and is appropriate for clinical data with asymmetric distributions.

Feature engineering was guided by the clinical understanding that cryopathy diagnosis depends not only on isolated laboratory values but also on their interactions. Seven derived features were therefore constructed. These included the interaction between cryoglobulin level and autohemagglutinin titer, the interaction between cryoglobulin level and isohemagglutinin titer, the ratio of autohemagglutinin to isohemagglutinin, the squared cryoglobulin level, the logarithmically transformed cryoglobulin level, the sum of the three titer measurements, and the number of positive binary tests. Together, these features were intended to capture non-linear effects, relative immunological dominance, and the cumulative burden of cold-reactive abnormalities. Categorical variables, including sex, age category, and examination period, were one-hot encoded.

At the normalization stage, all numerical features were standardized to zero mean and unit variance [36]. This ensured that variables measured on different scales contributed comparably to distance-based procedures such as Synthetic Minority Over-sampling Technique and to gradient-based optimisation in Gradient Boosted Trees and Multi-Layer Perceptron models. The final feature matrix contained 26 variables for 2,686 patients.

### *3.3 Classification Methods.*

All classification algorithms were implemented from scratch in NumPy without relying on external machine learning libraries, ensuring full algorithmic transparency and reproducibility. This design choice, although computationally more restrictive than optimized library-based implementations, provided full control over all modelling steps and avoided dependence on hidden defaults or undocumented heuristics. In the context of clinical decision support, such transparency is particularly important for methodological clarity and reproducibility.

#### *3.3.1 Decision Tree.*

The base classifier was a decision tree constructed according to the Classification and Regression Trees framework and using the Gini impurity criterion for split selection. At each internal node, a random subset of features was considered, and the splitting threshold was selected from percentile-based candidate values. This strategy served as a computationally efficient approximation to exhaustive threshold search. Tree growth was controlled by maximum depth, minimum leaf size, and node purity. To reduce memory overhead, trees were stored as compact nested tuples. The theoretical basis of this approach follows the classical CART framework introduced in [37].

#### *3.3.2 Random Forest.*

The Random Forest model followed the bagging framework introduced by Breiman and consisted of 40 to 50 independently trained decision trees. Each tree was fitted on a bootstrap sample of the training data, while random feature selection at each split reduced inter-tree correlation and improved ensemble diversity. Final predictions were obtained by majority voting. For probability estimation, which was required in soft-voting ensembles, class probabilities were computed as the proportion of trees voting for each class. The original Random Forest method was introduced in [38].

*3.3.3 Gradient Boosted Trees.*

The Gradient Boosted Trees model followed a multiclass softmax boosting scheme. The algorithm maintained a matrix of raw class scores that was iteratively updated. At each boosting step, class probabilities were computed using the softmax function, residuals were derived as the difference between the observed labels and predicted probabilities, and separate regression trees were fitted to these residuals. The updated score matrix was then adjusted using a learning-rate parameter. Four hyperparameter configurations were evaluated, differing in the number of iterations, tree depth, learning rate, and minimum leaf size. To improve efficiency, the regression trees used a reduced set of percentile-based split candidates. This formulation is grounded in Friedman's gradient boosting framework in [23].

*3.3.4 Multi-Layer Perceptron.*

The neural network model consisted of an input layer with 26 neurons, two hidden layers with Rectified Linear Unit activation, and a softmax output layer producing class probabilities for the 14 diagnostic categories. He initialization was used to improve training stability in the presence of rectified activations. Model training employed mini-batch stochastic gradient descent with cross-entropy loss, L2 regularization, and gradient clipping. Three network configurations were evaluated, differing in hidden layer size, regularization strength, learning rate, and number of epochs. No dropout, batch normalization, or learning-rate scheduling was applied, since the model was intended as a transparent neural baseline rather than a heavily optimized deep architecture. The use of rectified linear units is commonly linked to the work of Nair and Hinton [39]. The initialization scheme follows [40].

*3.3.5 Ensemble Methods.*

Three ensemble strategies were evaluated by combining base classifiers through soft voting, that is, by averaging their predicted class probabilities. This approach preserves uncertainty information that would otherwise be lost in hard majority voting. The evaluated variants included an equal-weight ensemble of Random Forest and Gradient Boosted Trees, a weighted version favouring Gradient

Boosted Trees, and a triple ensemble combining Random Forest, Gradient Boosted Trees, and the best-performing neural network. The theoretical rationale for combining classifier outputs through such fusion strategies is consistent with the framework proposed in [41].

*3.4 Class Balancing and Data Augmentation.*

The extreme class imbalance in the dataset posed a major challenge for multiclass classification, as standard models tend to favour the majority classes and may therefore achieve deceptively high overall accuracy while performing poorly on rare diagnoses. To address this issue, two balancing strategies were implemented.

The first strategy was Synthetic Minority Over-sampling Technique. Following the original formulation, synthetic minority-class samples were generated by interpolation between existing minority observations and their nearest neighbours in feature space. In the moderate balancing setting, minority classes were oversampled up to 70% of the majority class size. This allowed partial balancing while limiting the risk of generating an excessive number of ambiguous synthetic samples. The original SMOTE method was introduced in [42].

The second strategy was a more aggressive augmentation scheme combining full Synthetic Minority Over-sampling Technique balancing with additive Gaussian noise applied only to synthetic observations. The purpose of this step was to increase variability in the augmented data while avoiding major distortion of the class-conditional structure. This procedure expanded the training set substantially and allowed us to assess whether stronger balancing improved performance in the minority classes. The additional Gaussian perturbation was introduced here as an implementation-level augmentation step rather than as a direct replication of a single canonical method paper.

***3.5 Advanced Classification Strategies.***

*3.5.1 Hierarchical Classification.*

To reflect the clinical taxonomy of cryopathy syndromes, the 14-class problem was also reformulated as a two-level hierarchical task. At the first level, a Random Forest model predicted the broader pathogenetic category. At the second level, separate within-group classifiers were trained for those categories that contained more than one specific diagnosis. Final predictions were obtained by combining the predicted pathogenetic type with the diagnosis predicted inside that type. This strategy was motivated by the clinical assumption that broader mechanisms may be easier to distinguish than highly specific final diagnoses. Hierarchical decomposition is well established in machine learning and has also been discussed in clinically oriented decision-support settings, including recent medical informatics work [43].

#### *3.5.2 Binary Classifiers for Diagnostic Pairs.*

In addition to multiclass models, we trained dedicated binary classifiers for six clinically relevant diagnostic pairs. These pairs were selected on the basis of clinical similarity and confusion-matrix analysis. The underlying idea was that some differential diagnostic tasks may be more tractable when addressed directly as two-class problems rather than as part of a broader 14-class setting. Each binary classifier was implemented as a Random Forest model with class balancing adapted to the smaller diagnostic subset. The methodological basis of the classifier itself therefore follows the original Random Forest formulation of Breiman in [38], while the class-balancing step follows SMOTE [28].

#### *3.5.3 Period-Aware Models.*

To explore whether diagnostic patterns changed across epidemiological periods, separate Random Forest models were trained for the pre-COVID, COVID, and wartime subsets. Within each period, classes were remapped to match the locally observed label set, and balancing was performed separately. Predictions were then translated back to the original class indices for evaluation. This design allowed us to test whether period-specific modelling could better capture shifts in case mix, healthcare access, and disease presentation. From a methodological standpoint, these models again relied on Random Forest [38], together with Synthetic Minority Over-sampling Technique [28].

#### *3.5.4 Probability Calibration.*

Because raw ensemble probabilities are often poorly calibrated, Platt scaling was applied independently to each class. In this approach, the predicted probability for a given class is transformed through a logistic function with parameters estimated by minimizing binary cross-entropy loss. Calibration quality was evaluated using Expected Calibration Error on the calibration set, which measures the discrepancy between predicted probabilities and observed outcomes across probability bins. For the hold-out evaluation, probability calibration was performed using a dedicated calibration subset derived only from the training portion of the data. First, the full dataset was split into stratified training and test subsets. Next, the training subset was further divided into a model-fitting subset and a calibration subset. Base classifiers were trained only on the model-fitting subset, whereas the Platt scaling parameters were estimated exclusively on the calibration subset. The independent test subset was not used at any stage of probability calibration and was reserved solely for the final evaluation of calibrated predictions. This step was particularly important because the intended application involved clinical decision support rather than simple label assignment. Calibration quality was assessed using Expected

Calibration Error (ECE), a widely used metric for quantifying the discrepancy between predicted confidence and observed accuracy. Its modern formulation is discussed in [44] where ECE is explicitly defined and used as a primary calibration measure.

*3.6 Evaluation Protocol.*

Model evaluation was based on two complementary strategies: a fixed stratified train-test split for detailed comparative analysis and stratified five-fold cross-validation for robustness assessment.

In the train-test setting, the dataset was divided into 80% training and 20% test subsets using stratified sampling. This preserved the class distribution across both subsets and prevented the disappearance of rare classes from the test set. The same split was retained across all experiments to ensure fair comparison between methods.

For cross-validation, a stratified five-fold design was used. In each fold, class proportions were preserved and the full training pipeline, including balancing and model fitting, was performed independently. This prevented data leakage from the oversampling step and yielded more reliable performance estimates.

Given the strong imbalance and multiclass nature of the problem, evaluation relied on several complementary metrics. These included accuracy, macro-averaged F1 score, weighted F1 score, and Cohen's kappa [45]. In addition, per-class precision, recall, and F1 were analysed to identify which diagnoses were detected reliably and which remained difficult. Confusion matrices were used to examine systematic misclassification patterns. Expected Calibration Error was employed to assess the reliability of probability estimates, while Top-K accuracy was used to evaluate whether the correct diagnosis appeared among the model's top-ranked predictions. Finally, feature importance was estimated using permutation analysis, defined as the reduction in predictive performance after random shuffling of a given feature.

## Results

This section summarizes the main experimental results obtained for all modelling strategies examined in the study. The presentation moves from the overall comparison of methods to more focused analyses of ensemble behaviour, neural architectures, hierarchical modelling, binary diagnostic tasks, temporal robustness, probability calibration, and feature importance.

Because the dataset is strongly imbalanced and several diagnostic categories overlap clinically, the analysis gives particular weight to class-sensitive measures, especially macro F1 score and Cohen's kappa. Numerical results are reported first, whereas their broader interpretation is reserved for the discussion.

*4.1. Complete Model Comparison.*

Table 1. Overall comparison of all evaluated classification approaches ranked by macro F1 score.

| # | Approach | Accuracy | Macro F1 | Weighted F1 | Cohen's kappa | Notes |
|---|---|---|---|---|---|---|
| 1 | Ensemble: RF+GBT (equal) | 0.2664 | 0.1704 | 0.2520 | 0.1311 | Soft voting, 50/50 |
| 2 | Baseline RF + SMOTE | 0.2533 | 0.1685 | 0.2417 | 0.1198 | 50 trees, SMOTE 70% |
| 3 | R4: Period-Aware Models | 0.2720 | 0.1670 | 0.2487 | 0.1244 | 533/533 |
| 4 | Ensemble: RF+GBT (weighted) | 0.2664 | 0.1634 | 0.2501 | 0.1303 | Soft voting, 40/60 |
| 5 | Triple Ensemble (RF+GBT+MLP) | 0.2514 | 0.1596 | 0.2372 | 0.1220 | 3-model soft voting |
| 6 | R1: GBT Tuned + SMOTE | 0.3377 | 0.1496 | 0.2702 | 0.1654 | {'ni': 35, 'md': 4, 'lr': 0.15, 'ml': 8} |
| 7 | R2: Hierarchical (type to diagnosis) | 0.2495 | 0.1482 | 0.2383 | 0.1064 | L1 acc=0.386 |
| 8 | R5: RF + Aggressive Aug | 0.3809 | 0.1301 | 0.2746 | 0.1693 | SMOTE 100% + noise |
| 9 | MLP: MLP(128-64,reg+) | 0.1426 | 0.1122 | 0.1368 | 0.0631 | 3.5s |
| 10 | MLP: MLP(128-64) | 0.1351 | 0.1065 | 0.1268 | 0.0544 | 4.6s |
| 11 | MLP: MLP(64-32) | 0.1482 | 0.1015 | 0.1507 | 0.0479 | 2.6s |

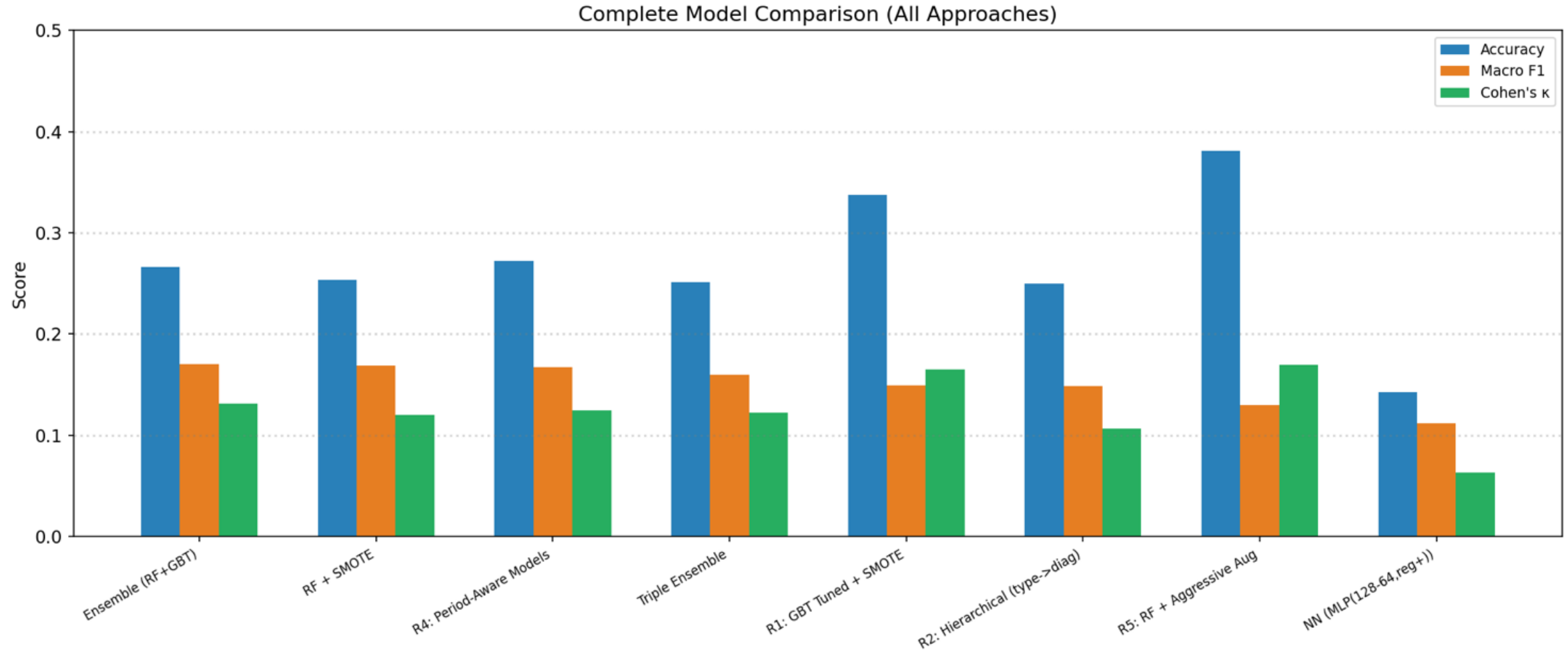

*Figure 1. Complete comparison of all classification approaches sorted by Macro F1.*

Table 1 presents the overall ranking of the evaluated approaches, ordered by macro F1 score.

Table 1 and Figure 1 summarize the overall ranking of the eleven evaluated approaches. The highest macro F1 score was obtained by the equal-weight ensemble combining Random Forest and Gradient Boosted Trees. By contrast, the weakest result was observed for the smallest Multi-Layer Perceptron configuration.

Even the strongest models remained in a relatively modest performance range, which underlines the difficulty of the task. The problem combines severe class imbalance with genuine clinical overlap between diagnostic categories, and these factors limit the attainable discrimination when only routine laboratory data are available.

A consistent trade-off between overall accuracy and macro F1 score was observed across the compared methods. Approaches based on stronger augmentation tended to improve majority-class recognition and therefore yielded higher accuracy, whereas balancing with Synthetic Minority Over-sampling Technique generally produced better minority-class sensitivity and, as a result, higher macro F1 values.

### *4.2 Ensemble Methods*

Table 2. Performance of individual and ensemble tree-based configurations.

| **Configuration** | **Accuracy** | **Macro F1** | **Weighted F1** | **Cohen's kappa** |
|---|---|---|---|---|
| RF alone | 0.2533 | 0.1685 | 0.2417 | 0.1198 |
| GBT alone | 0.3377 | 0.1496 | 0.2702 | 0.1654 |
| Ensemble equal | 0.2664 | 0.1704 | 0.2520 | 0.1311 |

| Ensemble weighted | 0.2664 | 0.1634 | 0.2501 | 0.1303 |
|---|---|---|---|---|
| Triple RF GBT NN | 0.2514 | 0.1596 | 0.2372 | 0.1220 |

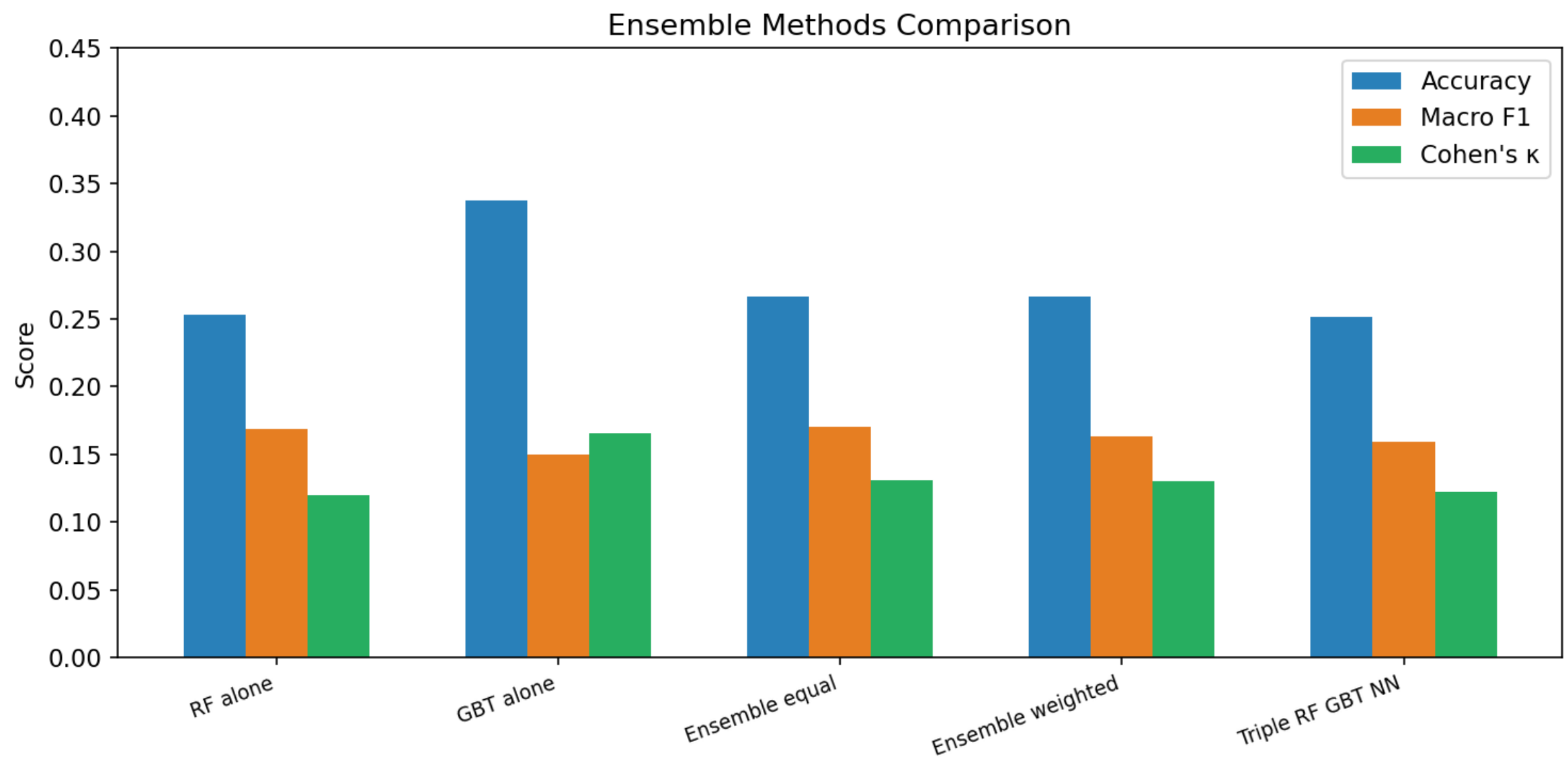

*Figure 2. Ensemble methods comparison.*

The ensemble analysis indicates that combining Random Forest with Gradient Boosted Trees is beneficial, although the gain remains moderate. The equal-weight ensemble achieved the best macro F1 score among the tested variants, which suggests that the two tree-based models contributed complementary information of comparable value.

Adding the neural network to the ensemble did not improve the final result. This suggests that, in the present setting, the neural model did not provide sufficiently informative probability estimates to strengthen the tree-based combination.

### *4.3 Neural Network Results*

Table 3. Performance of the evaluated neural network architectures.

| **Architecture** | **Accuracy** | **Macro F1** | **Weighted F1** | **Cohen's kappa** | **Time** |
|---|---|---|---|---|---|
| MLP(64-32) | 0.1482 | 0.1015 | 0.1507 | 0.0479 | 2.6s |
| MLP(128-64) | 0.1351 | 0.1065 | 0.1268 | 0.0544 | 4.6s |
| MLP(128-64,reg+) | 0.1426 | 0.1122 | 0.1368 | 0.0631 | 3.5s |

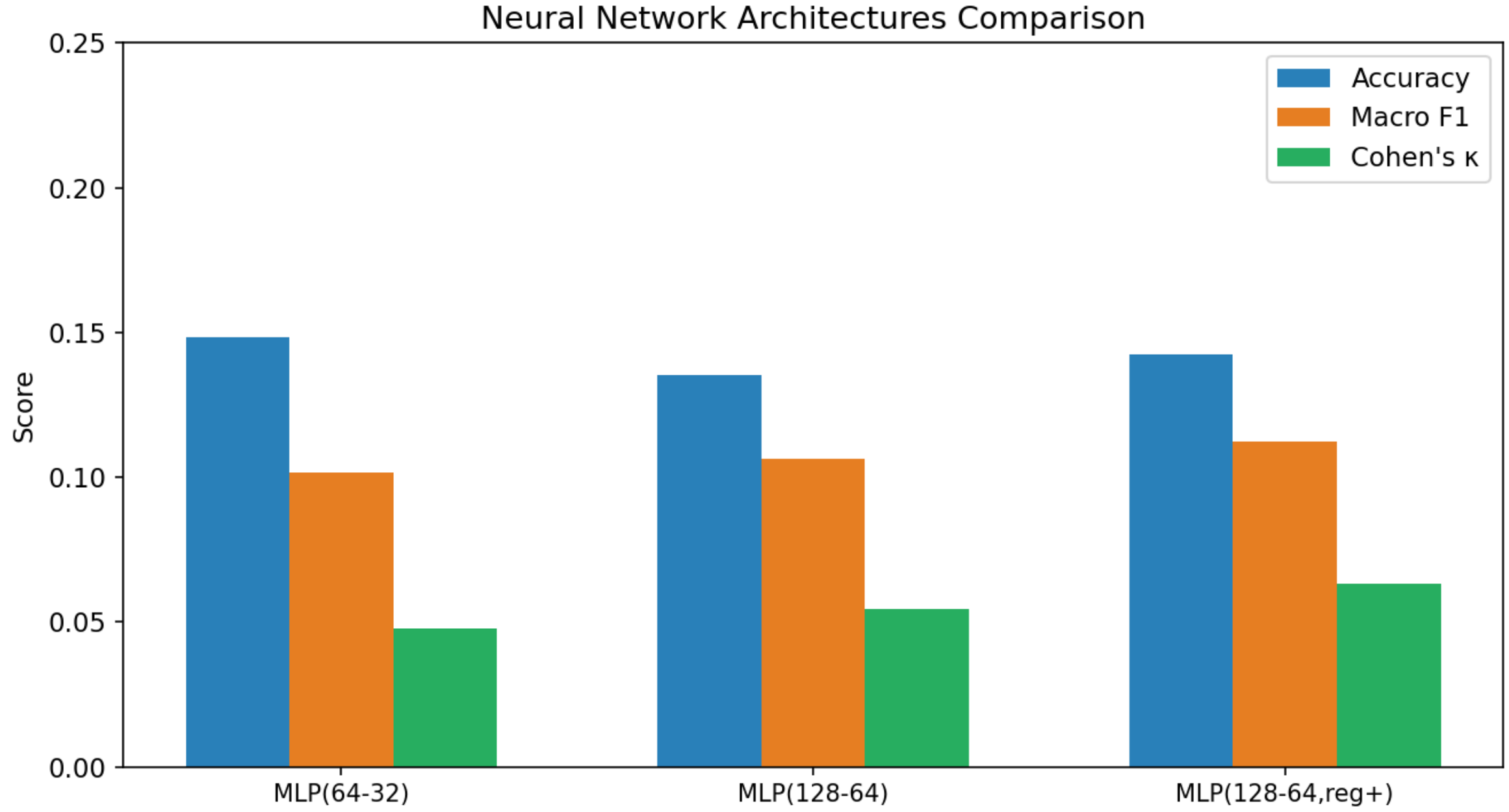

*Figure 3. Neural network architectures comparison.*

All three Multi-Layer Perceptron configurations performed worse than the tree-based methods. Even the strongest neural model remained clearly below the balanced Random Forest baseline in both macro F1 score and overall agreement.

The differences between the tested network sizes were small. This suggests that the main limitation was not model capacity itself, but the restricted amount of informative training signal available across fourteen classes.

Increasing regularization also led to only minor change. Taken together, these results support the view that, for this type of small and heterogeneous tabular clinical dataset, tree-based ensembles remain a more appropriate choice than simple feed-forward neural networks.

### *4.4 Gradient Boosted Trees Hyperparameter Tuning*

Table 4. Results of Gradient Boosted Trees hyperparameter tuning.

| # | Config | Accuracy | Macro F1 | Cohen's kappa |
|---|---|---|---|---|
| 1 | {'ni': 25, 'md': 3, 'lr': 0.1, 'ml': 8} | 0.3621 | 0.1369 | 0.1559 |
| 2 | {'ni': 35, 'md': 4, 'lr': 0.15, 'ml': 8} | 0.3377 | 0.1496 | 0.1654 |
| 3 | {'ni': 30, 'md': 3, 'lr': 0.2, 'ml': 10} | 0.3246 | 0.1453 | 0.1560 |
| 4 | {'ni': 35, 'md': 3, 'lr': 0.15, 'ml': 5} | 0.3283 | 0.1412 | 0.1504 |

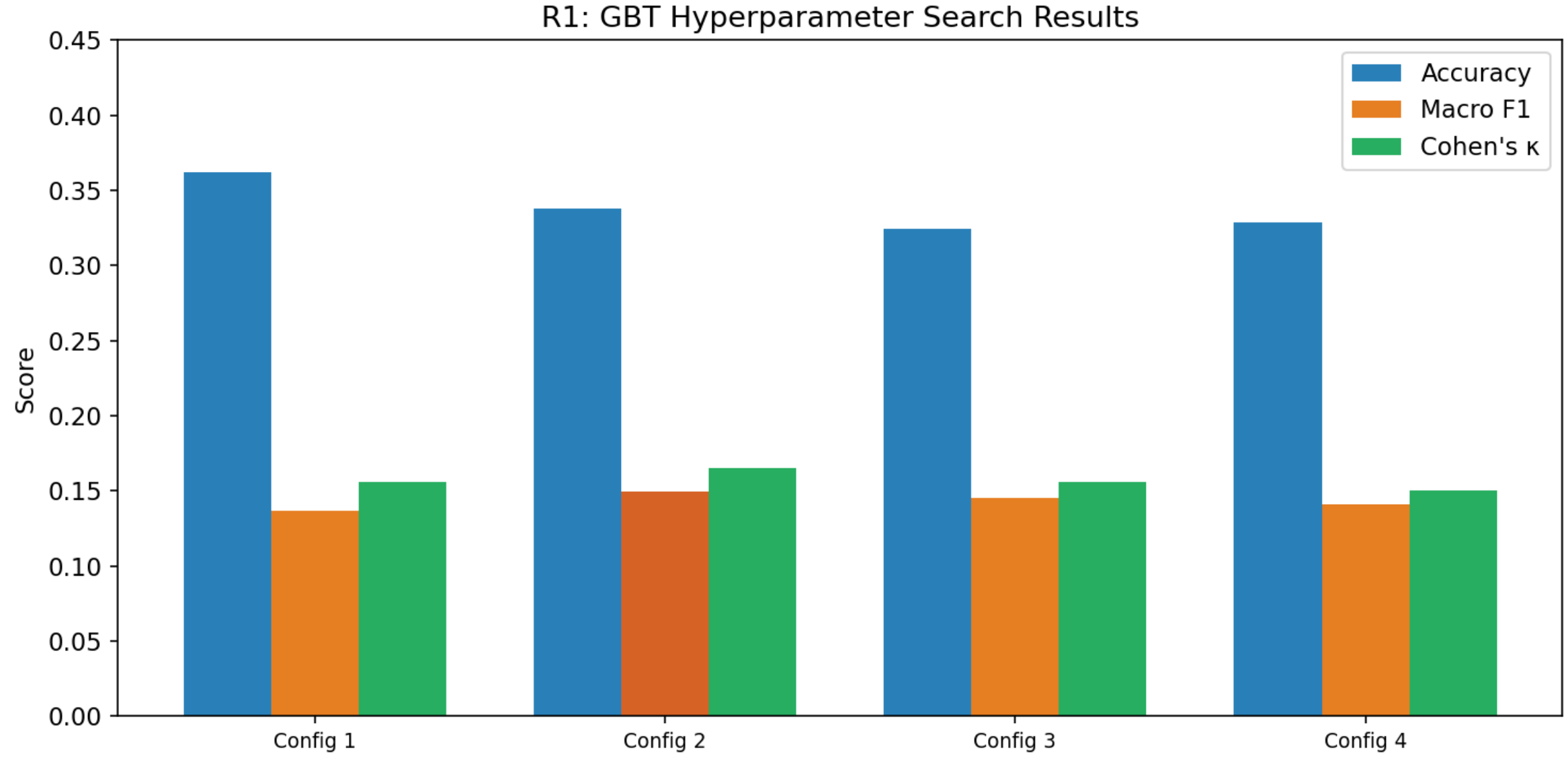

*Figure 4. GBT hyperparameter search results.*

The hyperparameter search showed that Gradient Boosted Trees were relatively stable across the tested configurations. The spread in macro F1 score between the best and weakest setting was small, which indicates that model behaviour was not overly sensitive to moderate changes in the main tuning parameters.

The best result was obtained with a moderate learning rate and deeper trees. At the same time, the differences between the tested settings were not large, which suggests that the discriminative structure of the feature space is only moderately affected by further tuning within this range.

### *4.5 Hierarchical Classification*

Table 5. Performance of the hierarchical classification strategy.

| **Level** | **Accuracy** | **Macro F1** | **Cohen's kappa** | **Details** |
|---|---|---|---|---|
| L1: Pathogenetic Type (7 classes) | 0.3865 | 0.2655 | 0.1398 | RF 50 trees + SMOTE |
| Combined L1 to L2 (14 classes) | 0.2495 | 0.1482 | 0.1064 | Sub-classifiers per type |

The hierarchical strategy produced a mixed result. At the level of broader pathogenetic categories, the model separated the classes more successfully than in the full fourteen-class setting, which indicates that the higher-level diagnostic structure is indeed easier to learn.

However, the complete two-stage pipeline did not outperform the direct multiclass baseline. The most plausible explanation is error propagation: once a case is assigned to the wrong broad category at the first level, the second stage cannot recover the correct final diagnosis.

*4.6 Binary Classifiers for Key Diagnostic Pairs*

Table 6. Results of binary classifiers for clinically relevant diagnostic pairs.

| Pair | Accuracy | F1 | Cohen's kappa | n | Agreement |
|---|---|---|---|---|---|
| Cryovasculitis vs IDD | 0.7312 | 0.7126 | 0.4607 | 279 | Substantial |
| Cold Allergy vs Raynaud | 0.6098 | 0.3846 | 0.1206 | 82 | Slight |
| SLE vs Systemic Vasculitis | 0.5571 | 0.3673 | 0.0734 | 70 | Slight |
| Cryovasculitis vs Systemic Vasculitis | 0.6358 | 0.7586 | 0.0210 | 173 | Slight |
| Hepatitis C vs Chronic Hepatitis | 0.7895 | 0.8000 | 0.5824 | 19 | Substantial |
| Rheumatoid vs Reactive Arthritis | 0.5128 | 0.1739 | -0.1488 | 39 | Poor |

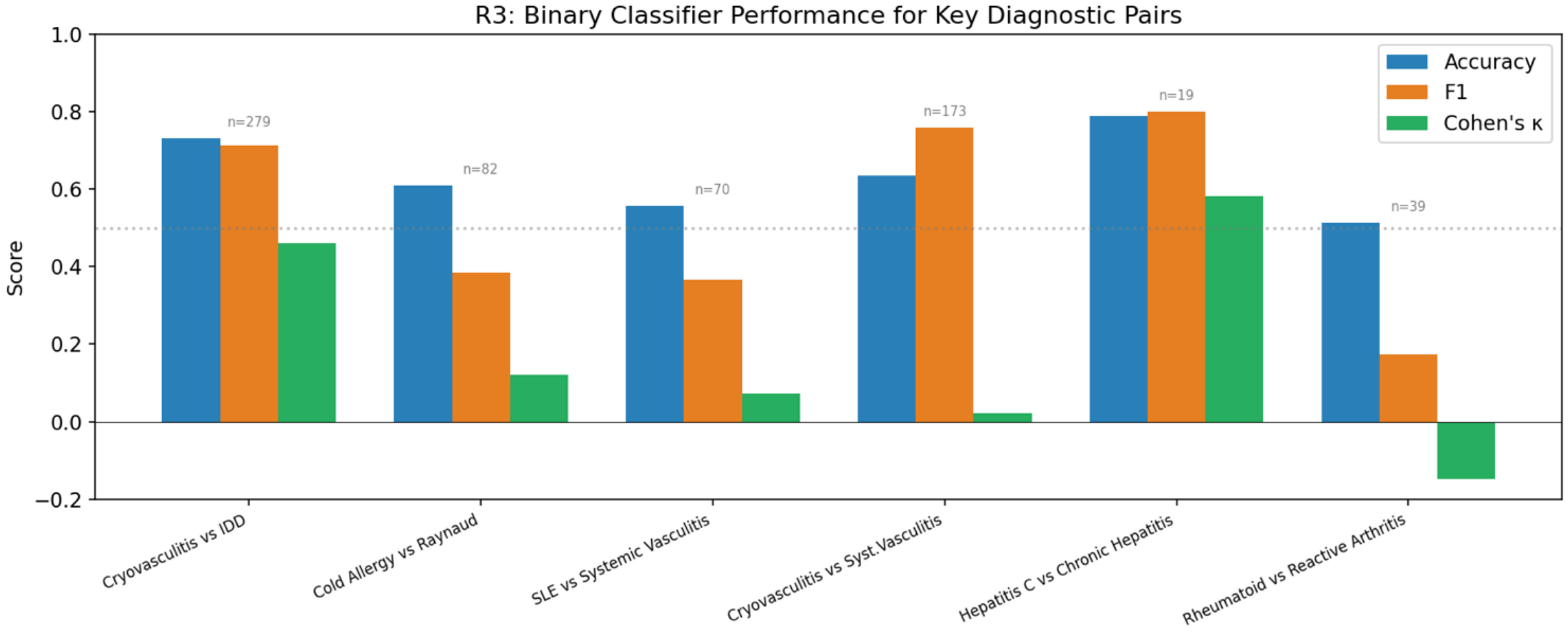

*Figure 5. Binary classifier performance for diagnostic pairs.*

The binary experiments yielded the strongest results in the study. For selected differential diagnosis pairs, the models reached a level of performance that is substantially higher than that observed in the full multiclass setting.

The clearest result was obtained for Hepatitis C versus Chronic Hepatitis, while Cryovasculitis versus immunodeficiency-related disease also showed practically meaningful discrimination. These findings suggest that targeted binary support may be especially valuable in clinically frequent or diagnostically difficult pairwise scenarios.

At the same time, not all pairs were equally tractable. Pairs involving closely related autoimmune conditions remained difficult, which is consistent with the underlying serological overlap and with the clinical complexity of these distinctions.

### *4.7 Period-Aware Models*

Table 7. Results of period-aware modelling across epidemiological subsets.

| Period | Accuracy | Macro F1 | Cohen's kappa | Classes | n |
|---|---|---|---|---|---|
| pre-COVID | 0.2845 | 0.1660 | 0.1420 | 14 | 355 |
| during COVID | 0.2632 | 0.1092 | 0.0859 | 12 | 19 |
| wartime | 0.2484 | 0.1419 | 0.0992 | 13 | 157 |
| Combined | 0.2720 | 0.1670 | 0.1244 | — | 533/533 |

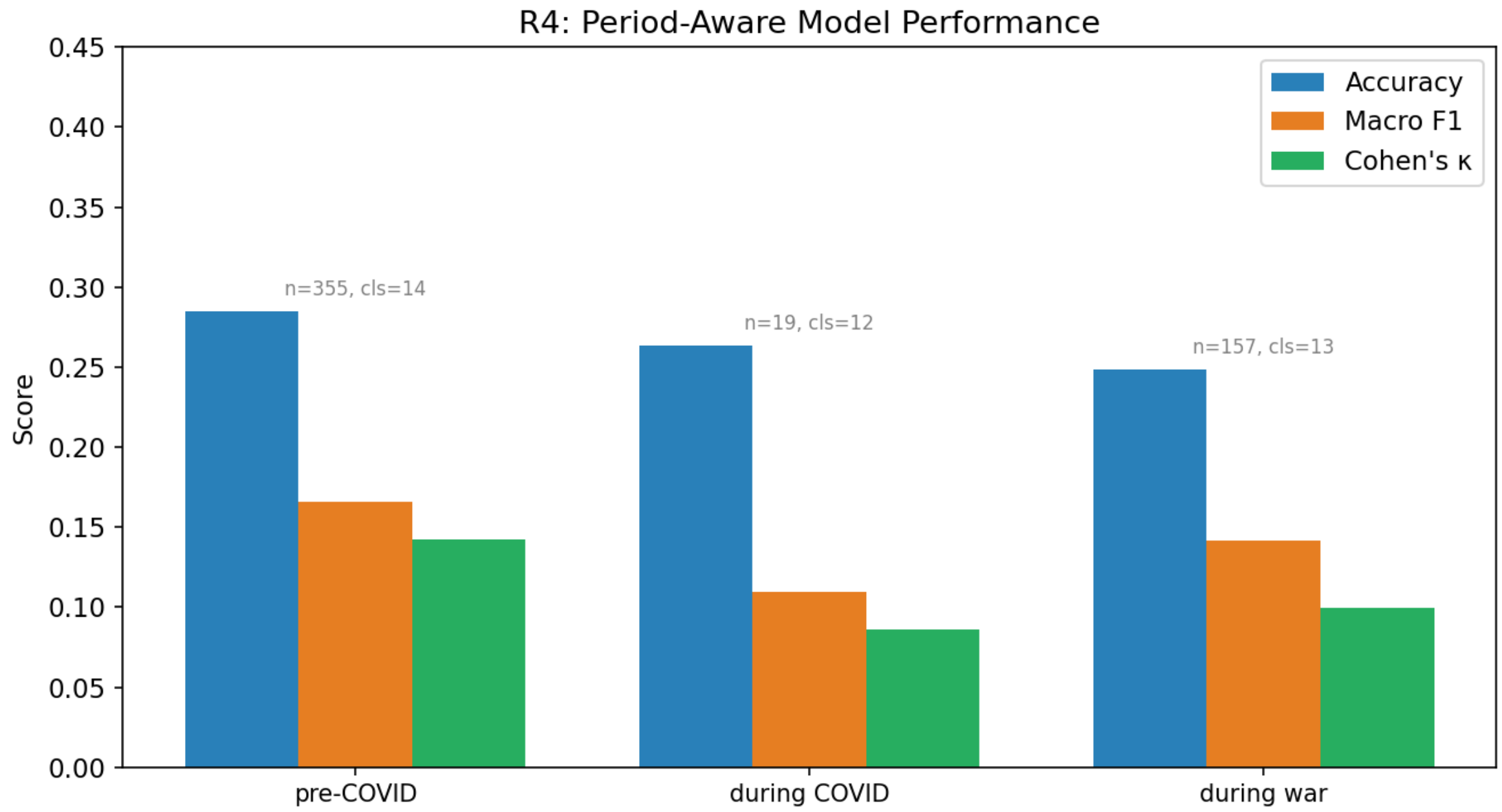

*Figure 6. Period-aware model performance.*

The period-aware analysis did not show a clear advantage for training separate models for the pre-COVID, COVID, and wartime subsets. The combined system reached performance close to that of the global model, but not above it.

This finding suggests that the relationship between laboratory variables and diagnosis remained broadly stable across the examined epidemiological periods. From a practical perspective, this is encouraging, as it indicates that models trained on historical data may retain relevance under changing clinical conditions.

*4.8 Cross-Validation*

Table 8. Stratified five-fold cross-validation results.

| **Model** | **Metric** | **Mean** | **Std** | **F1** | **F2** | **F3** | **F4** | **F5** |
|---|---|---|---|---|---|---|---|---|
| RF | Accuracy | 0.3857 | 0.0052 | 0.3775 | 0.3833 | 0.3881 | 0.3933 | 0.3865 |
| RF | Macro F1 | 0.1231 | 0.0188 | 0.1448 | 0.1057 | 0.0963 | 0.1316 | 0.1372 |
| RF | Cohen's κ | 0.1782 | 0.0061 | 0.1722 | 0.1762 | 0.1801 | 0.1891 | 0.1736 |
| RF + SMOTE | Accuracy | 0.2483 | 0.0185 | 0.2578 | 0.2352 | 0.2743 | 0.2210 | 0.2533 |
| RF + SMOTE | Macro F1 | 0.1571 | 0.0143 | 0.1801 | 0.1429 | 0.1415 | 0.1642 | 0.1570 |
| RF + SMOTE | Cohen's κ | 0.1283 | 0.0161 | 0.1378 | 0.1149 | 0.1491 | 0.1048 | 0.1349 |

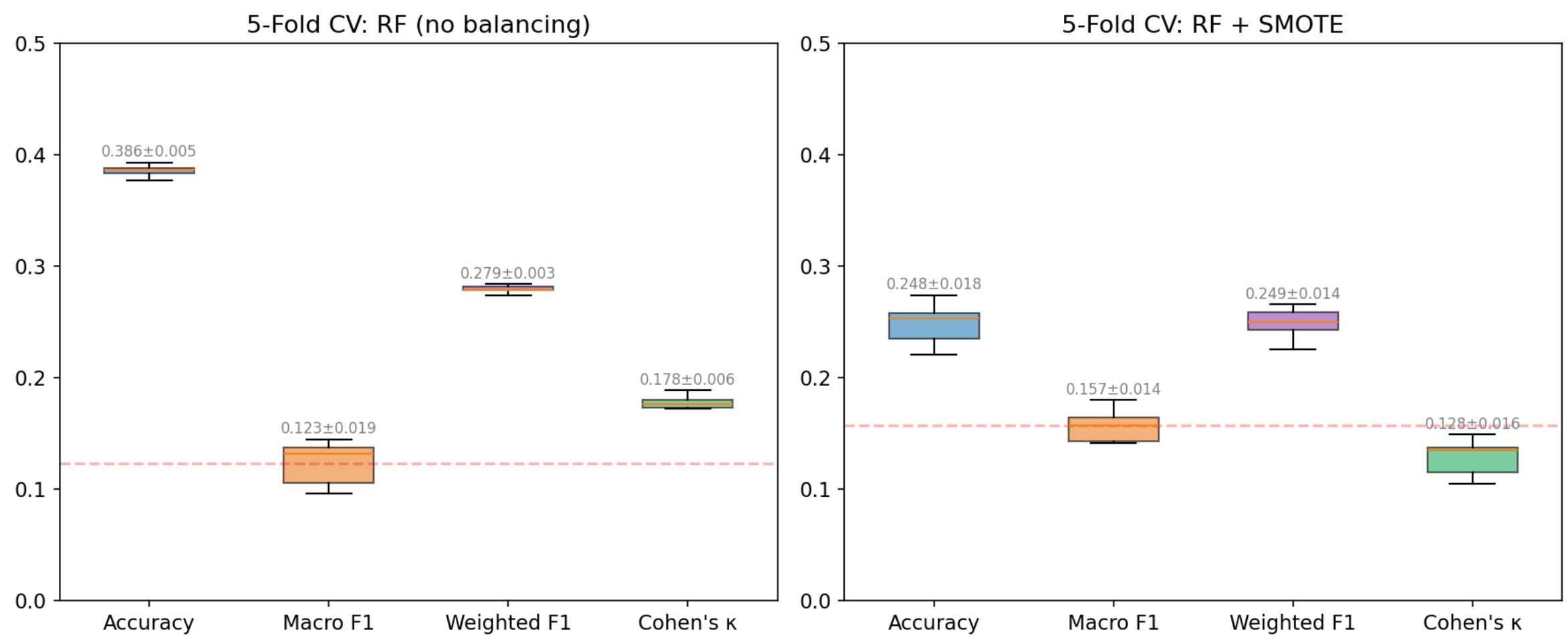

*Figure 7. Cross-validation distributions (5-fold stratified).*

Five-fold stratified cross-validation confirmed that the main observations were not driven by a favourable single split of the data. In particular, the contrast between the unbalanced Random Forest and the balanced Random Forest with Synthetic Minority Over-sampling Technique was reproduced consistently across folds.

The standard deviations were low, which supports the robustness of the reported tendencies. In particular, the trade-off between higher accuracy for the unbalanced model and higher macro F1 score for the balanced model was stable across repeated partitions.

*4.9 Probability Calibration*

Table 9. Class-wise probability calibration before and after Platt scaling.

| **Class** | **ECE (Raw)** | **ECE (Calibrated)** | **Improvement** | **Support** |
|---|---|---|---|---|
| IDD | 0.1404 | 0.0038 | +97.3% | 144 |
| Cryovasculitis | 0.1209 | 0.1068 | +11.6% | 135 |
| Raynaud | 0.0360 | 0.0204 | +43.3% | 50 |
| Systemic Vasculitis | 0.0202 | 0.0259 | -28.4% | 38 |
| SLE | 0.0202 | 0.0290 | -43.9% | 32 |
| Cold Allergy | 0.0110 | 0.0291 | -165.6% | 32 |
| Reactive Arthritis | 0.0179 | 0.0332 | -85.5% | 25 |
| Scleroderma | 0.0324 | 0.0399 | -23.3% | 15 |
| Rheumatoid Arthritis | 0.0352 | 0.0407 | -15.6% | 14 |
| Other | 0.0259 | 0.0423 | -63.2% | 12 |
| Hepatitis C | 0.0569 | 0.0429 | +24.5% | 11 |
| Control | 0.0376 | 0.0448 | -19.1% | 9 |
| Hypothyroidism | 0.0425 | 0.0454 | -6.8% | 8 |
| Chronic Hepatitis | 0.0407 | 0.0454 | -11.5% | 8 |

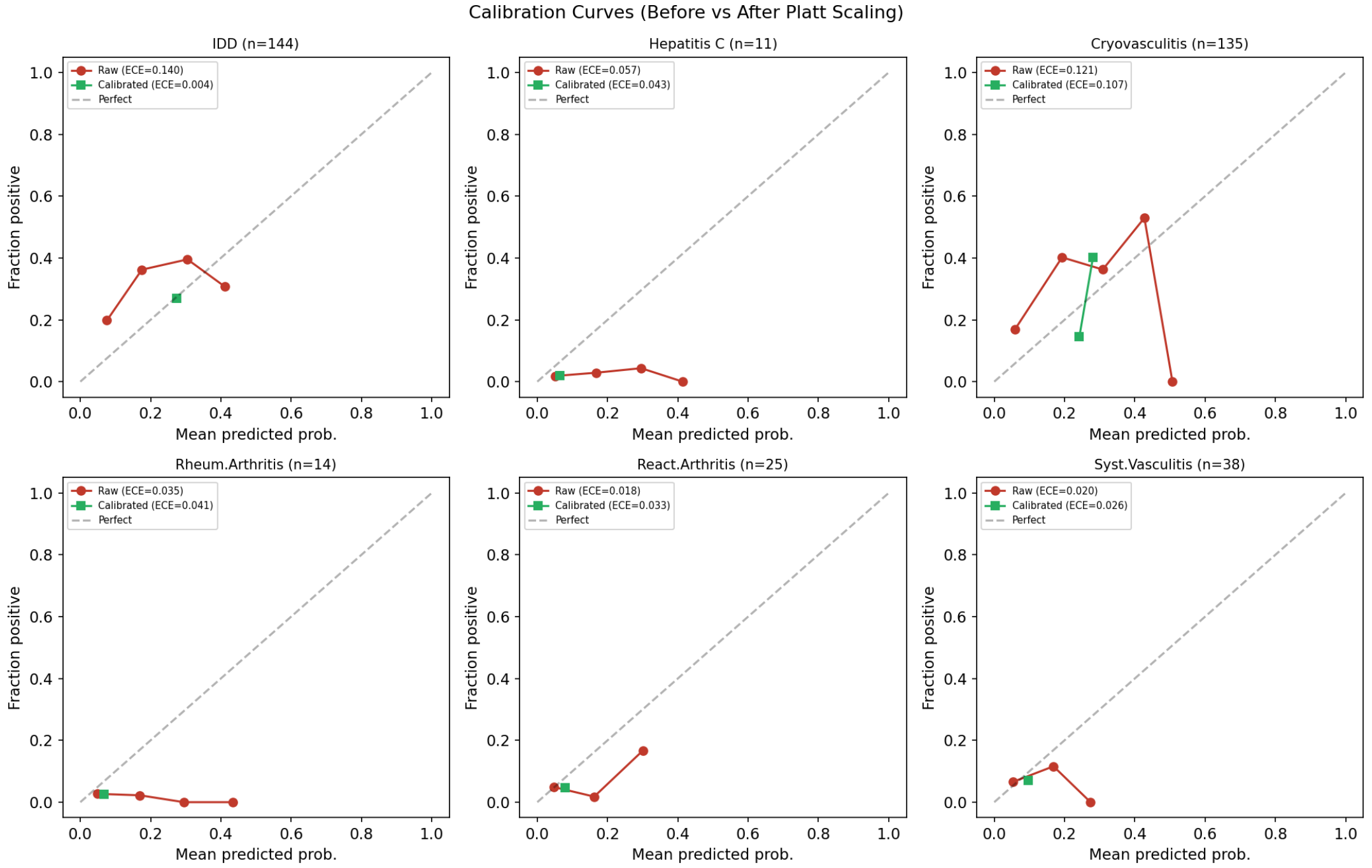

*Figure 8. Calibration curves before and after Platt scaling.*

Probability calibration produced a selective rather than universal benefit. For several classes, especially the largest one, calibration reduced the discrepancy between predicted and observed probabilities in a substantial way.

For smaller classes, the effect was less consistent and in some cases adverse. This is not unexpected, because calibration becomes difficult when only a small number of positive examples is available for estimating the transformation reliably.

### *4.10 Top-K Accuracy and Confidence*

Table 10. Top-K accuracy and prediction confidence.

| **Metric** | **Value** |
|---|---|
| Top-1 Accuracy | 0.2664 |
| Top-2 Accuracy | 0.3921 |
| Top-3 Accuracy | 0.5310 |
| Confidence on correct predictions | 0.2956 |
| Confidence on incorrect predictions | 0.2622 |

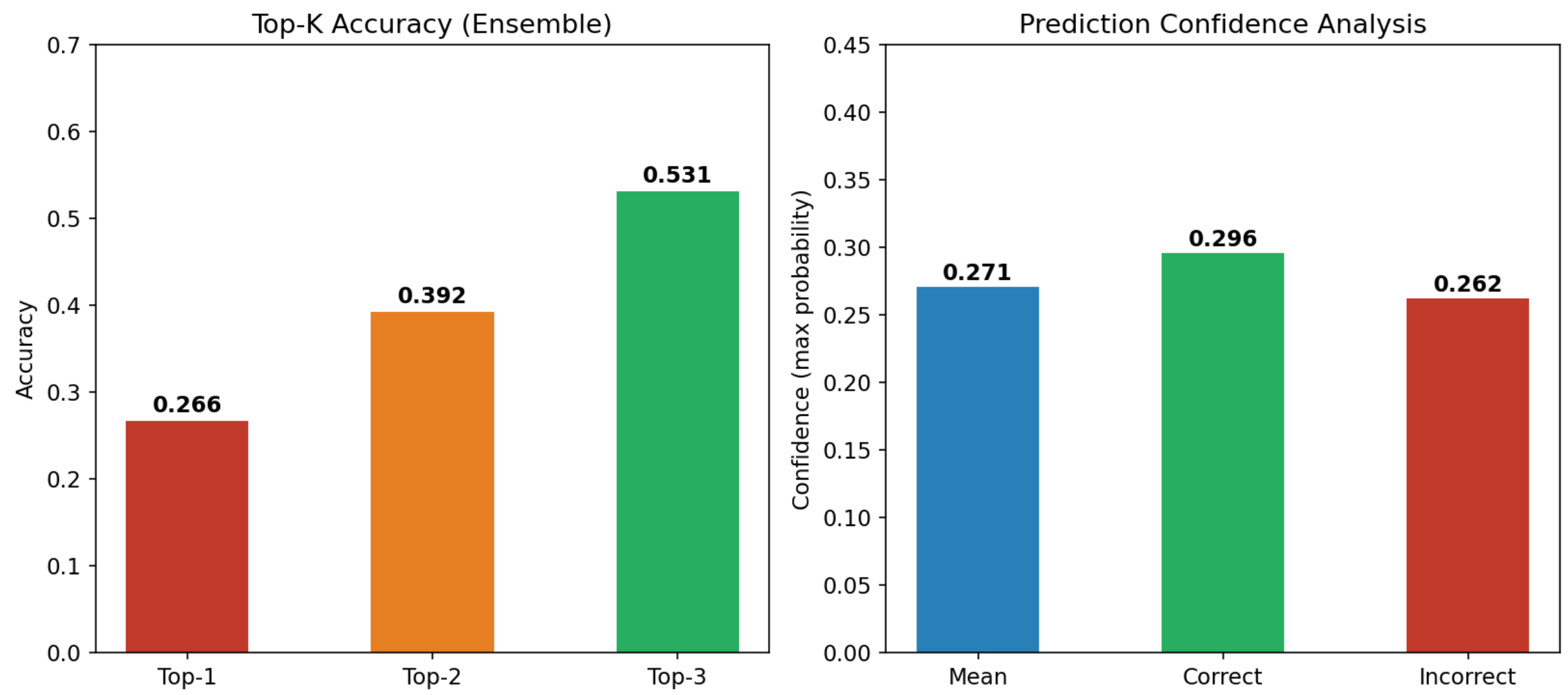

*Figure 9. Top-K accuracy and confidence analysis.*

The Top-K analysis provides one of the most practically relevant findings of the study. Although the top-1 accuracy remained limited, the correct diagnosis appeared among the three highest-ranked predictions in more than half of the cases.

This suggests that the proposed system may be more useful as a screening or shortlisting tool than as a fully autonomous classifier. In clinical use, presenting a small set of plausible candidates together with their estimated probabilities may offer a more realistic form of support.

### *4.11 Per-Class Performance*

Table 11. Per-class precision, recall, and F1 score for the best ensemble model.

| **Diagnosis** | **Precision** | **Recall** | **F1** | **Support** |
|---|---|---|---|---|
| IDD | 0.341 | 0.326 | 0.333 | 144 |
| Hepatitis C | 0.025 | 0.091 | 0.039 | 11 |
| Hypothyroidism | 0.129 | 0.500 | 0.205 | 8 |
| Control | 0.286 | 0.889 | 0.432 | 9 |
| Cryovasculitis | 0.449 | 0.489 | 0.468 | 135 |
| Rheumatoid Arthritis | 0.048 | 0.071 | 0.057 | 14 |
| Reactive Arthritis | 0.053 | 0.040 | 0.045 | 25 |
| Systemic Vasculitis | 0.000 | 0.000 | 0.000 | 38 |
| SLE | 0.125 | 0.062 | 0.083 | 32 |
| Scleroderma | 0.043 | 0.067 | 0.053 | 15 |
| Cold Allergy | 0.167 | 0.156 | 0.161 | 32 |
| Chronic Hepatitis | 0.167 | 0.250 | 0.200 | 8 |
| Raynaud | 0.125 | 0.020 | 0.034 | 50 |
| Other | 0.300 | 0.250 | 0.273 | 12 |

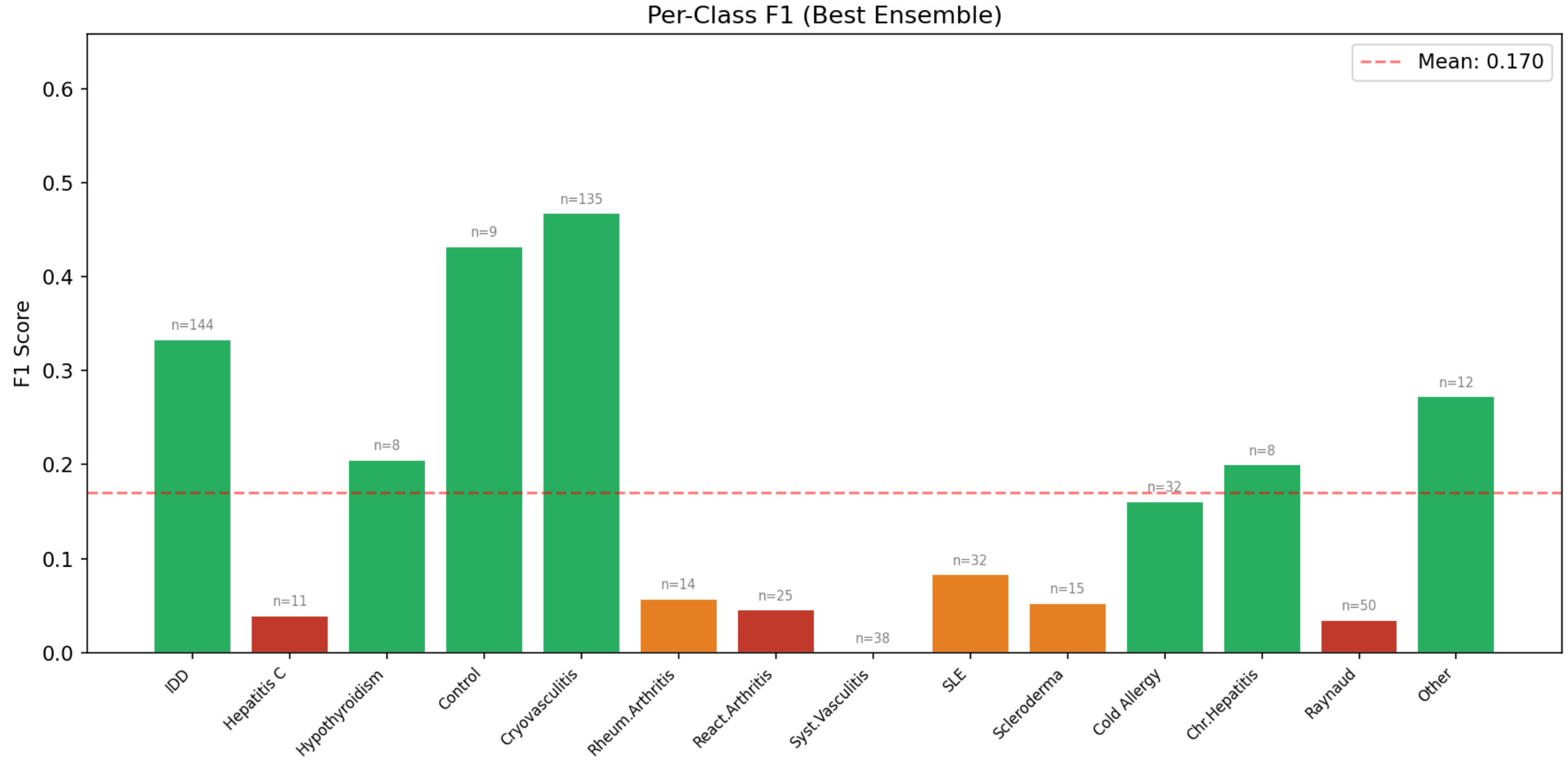

*Figure 10. Per-class F1 scores with sample sizes annotated.*

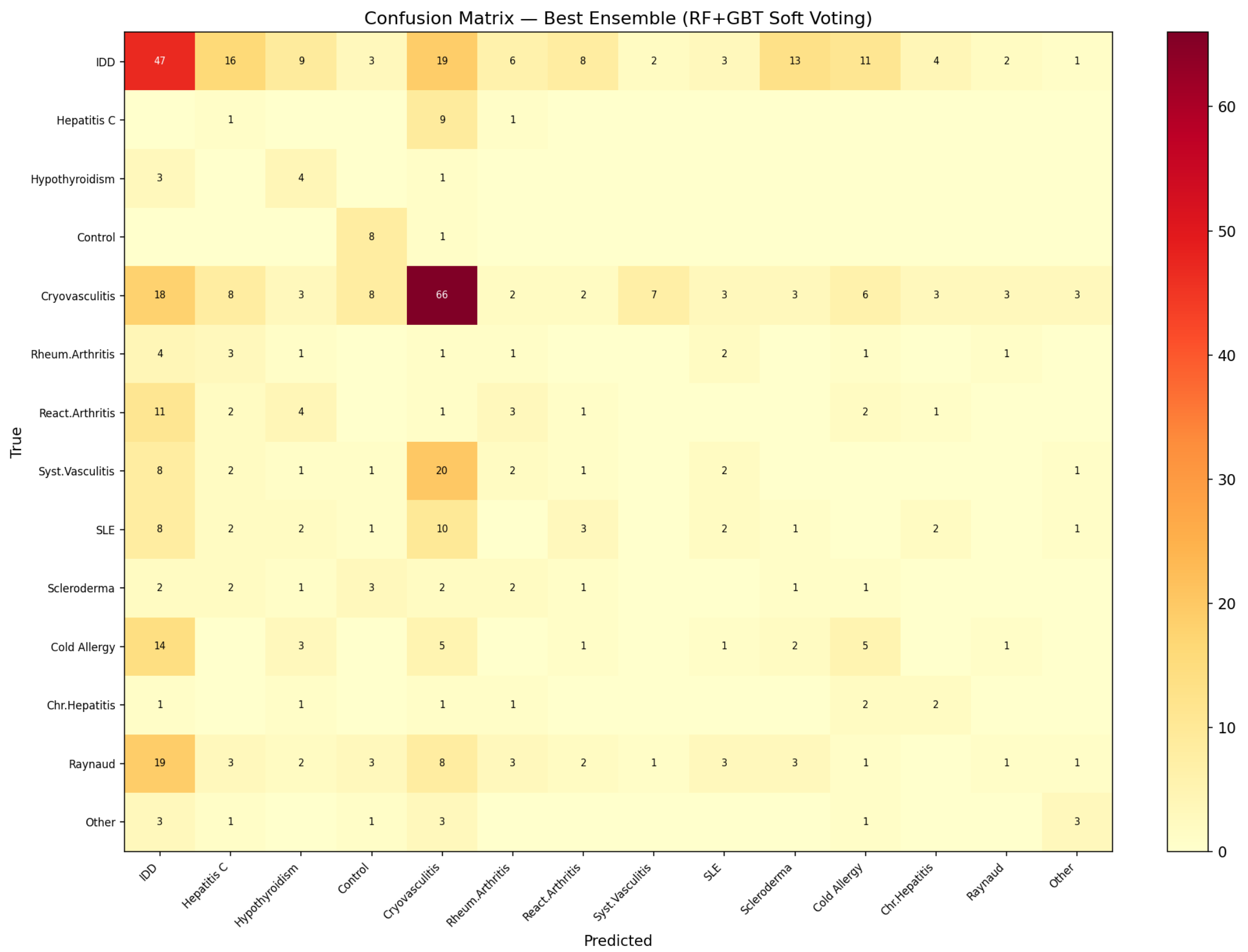

*Figure 11. Confusion matrix (best ensemble).*

The per-class analysis revealed marked heterogeneity in model behaviour. A small number of classes, particularly Cryovasculitis and the control group, achieved clearly better discrimination, whereas several minority diagnoses remained difficult to detect.

This pattern shows that the difficulty of the problem is not evenly distributed across classes. Performance depends not only on sample size, but also on how distinct the corresponding laboratory profile is relative to the rest of the diagnostic spectrum.

### *4.12 Feature Importance*

Table 12. Permutation-based feature importance for the best ensemble model.

| **Rank** | **Feature** | **Importance (Δ Accuracy)** |
|---|---|---|
| 1 | Log-transformed cryoglobulin level | 0.0263 |
| 2 | Cryoglobulin × Autohemagglutinin titer | 0.0206 |
| 3 | Squared cryoglobulin level | 0.0206 |
| 4 | Auto/Iso hemagglutinin ratio | 0.0188 |
| 5 | Cryoglobulin × Isohemagglutinin titer | 0.0150 |
| 6 | Spectrophotometric cryoglobulin level | 0.0094 |
| 7 | Cold isohemagglutinin titer | 0.0075 |
| 8 | Cold autohemagglutinin titer | 0.0056 |
| 9 | Total titer sum | 0.0056 |
| 10 | Number of positive tests | 0.0056 |
| 11 | Sex: Female | 0.0038 |
| 12 | Age category: 45-59 | 0.0038 |
| 13 | Age | 0.0019 |
| 14 | Sia test: paraproteins | 0.0019 |
| 15 | Sia test: macroglobulins | 0.0019 |
| 16 | Cold plasma precipitate | 0.0019 |
| 17 | True cold precipitation | 0.0000 |
| 18 | Biphasic hemolysin titer | 0.0000 |
| 19 | Age category: 30-44 | 0.0000 |
| 20 | Period: during COVID | -0.0019 |

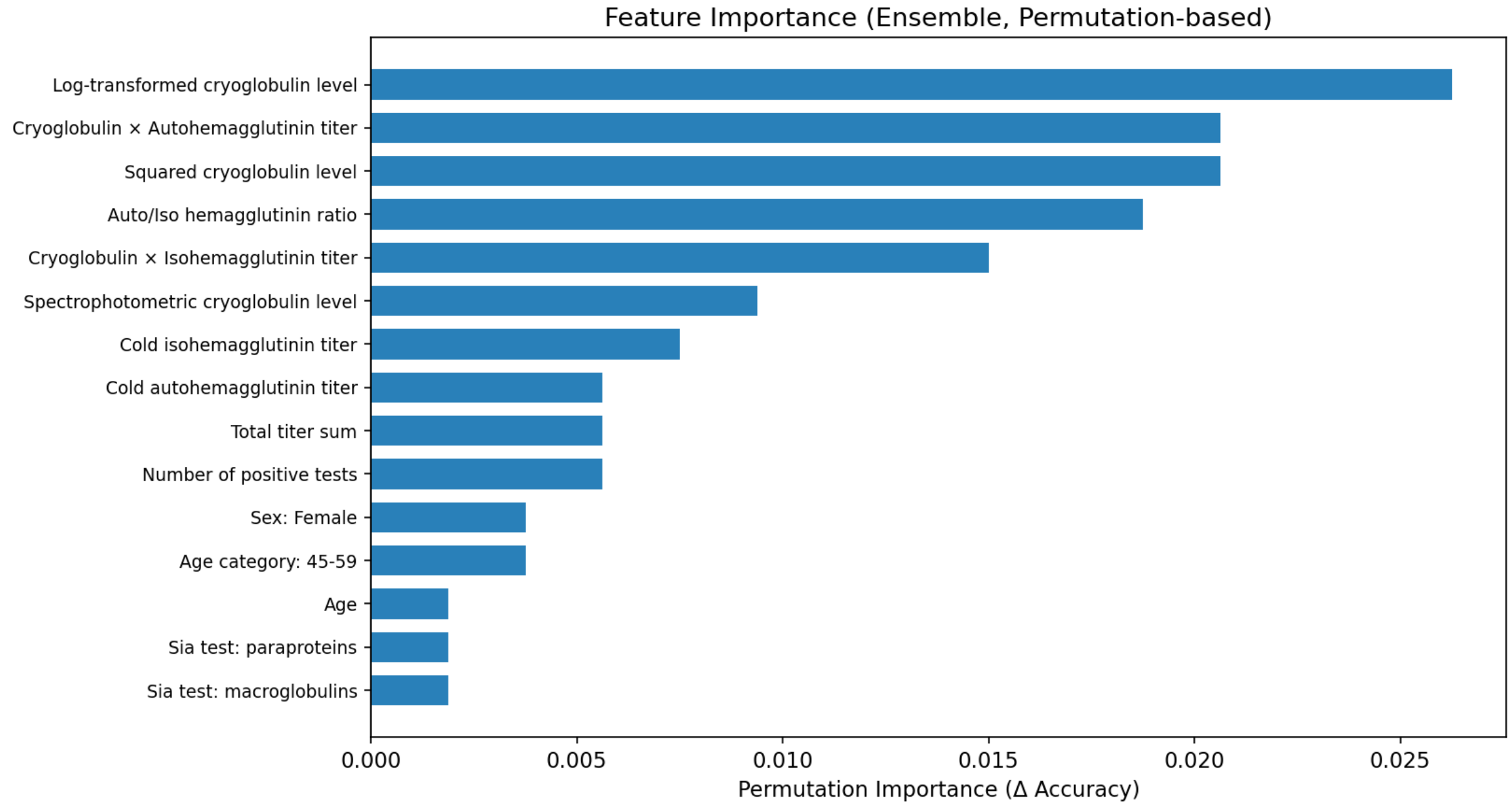

*Figure 12. Permutation-based feature importance (top 15 features).*

Permutation analysis showed that the most informative predictors were the cryoglobulin-derived variables and their engineered transformations. In particular, the logarithmically transformed cryoglobulin level and the interaction terms involving hemagglutinin titers ranked among the strongest features.

These results support the clinical rationale behind the feature engineering stage. They also suggest that the diagnostic signal is driven primarily by quantitative laboratory patterns rather than by demographic characteristics or examination period alone.

## Discussion

The present study examined the feasibility of machine learning-based classification of cryopathy syndromes in a highly imbalanced multiclass setting. The results show that this is a difficult diagnostic problem, both computationally and clinically. At the same time, they demonstrate that meaningful structure can still be extracted from routine cryoglobulin-related laboratory data, especially when the modelling strategy is aligned with the characteristics of the task.

The strongest overall performance was achieved by the soft-voting ensemble that combined Random Forest and Gradient Boosted Trees. Although the absolute values of the main performance measures remained moderate, this result should be interpreted in the context of the problem itself. The task involved 14 diagnostic categories, substantial overlap between clinical groups, a limited set of laboratory variables, and a pronounced class imbalance. Under such conditions, even modest gains in macro F1 score are informative, because they reflect improved recognition of minority diagnostic classes rather than better performance only on the dominant categories.

One of the clearest findings of the study is the persistent trade-off between overall accuracy and class-sensitive performance. Approaches based on stronger augmentation tended to improve accuracy, but this gain was accompanied by a reduction in macro F1 score. In practical terms, this means that the model became better at predicting frequent diagnoses while losing sensitivity to rarer categories. By contrast, balancing strategies based on Synthetic Minority Over-sampling Technique improved performance in terms of macro F1 score, although at the cost of lower overall accuracy. This pattern was observed both in the hold-out evaluation and in cross-validation, which suggests that it reflects a genuine property of the data. For this reason, model quality in this setting should not be judged by accuracy alone.

The comparison of modelling approaches also supports the use of tree-based ensembles as the most appropriate baseline for this type of data. Neural networks consistently underperformed both Random Forest and Gradient Boosted Trees. This difference is not surprising. The dataset is relatively small for deep learning, the features are heterogeneous, and the decision boundaries appear to be irregular rather than smooth. In such settings, tree-based models usually make better use of the available signal. The present results therefore align with broader evidence that, for small and medium-sized tabular medical datasets, ensemble methods often remain more reliable than neural architectures.

Another important observation concerns the value of feature engineering. Several of the most informative predictors were not raw laboratory measurements but derived variables reflecting interactions between cryoglobulin level and hemagglutinin-related indicators. This suggests that clinically informed transformations can help expose structure that is less visible in the original feature space. In other words, the improvement did not come only from model choice; it also depended on how the biological meaning of the data was encoded before training. This is particularly relevant for medical machine learning, where the combination of computational methods with domain knowledge often determines whether a model becomes merely predictive or genuinely useful.

From a clinical perspective, perhaps the most encouraging result is that targeted binary classifiers performed substantially better than the full multiclass models. In particular, the distinction between Hepatitis C and Chronic Hepatitis, as well as between Cryovasculitis and immunodeficiency-related disease, reached performance levels that may be meaningful for decision support. This suggests that the most practical use of the proposed framework may not be fully automated 14-class diagnosis. A more realistic application would be staged support: first, a multiclass model generates a short list of the most plausible diagnoses; then, dedicated binary models assist with specific differential decisions in clinically difficult pairs. Such an approach is more consistent with real diagnostic reasoning and is more likely to be accepted in practice.

The Top-3 analysis further supports this interpretation. While Top-1 performance remained limited, the correct diagnosis appeared within the three highest-ranked model outputs in more than half of the cases. This finding is important because, in many clinical settings, decision support is not expected to replace specialist judgement. Its value lies in narrowing the search space, highlighting plausible alternatives, and reducing diagnostic oversight. In this sense, a shortlist-based system may be more useful than a single-label predictor, especially when laboratory information alone is insufficient for definitive discrimination.

Probability calibration adds another practical dimension to the results. For the largest diagnostic classes, calibration improved noticeably after Platt scaling, which means that the predicted probabilities became more interpretable as confidence estimates. This matters in clinical use, because physicians need to understand not only which diagnosis is ranked first, but also how certain or uncertain the model is. At the same time, the calibration results also show an important limitation: improvement was not uniform across all classes, and the smallest groups remained difficult to calibrate reliably. This is expected when support is limited, but it also indicates that probability-based outputs should be interpreted with greater caution in rare diagnoses.

The hierarchical classification experiment provided a useful methodological insight. At the broader pathogenetic level, discrimination was substantially better than in the final 14-class setting. This suggests that the structure of the problem is indeed hierarchical and that the broader biological grouping is easier to recover from the available data. However, the full two-stage cascade did not outperform the direct multiclass baseline. Most likely, this reflects error propagation from the first stage to the second, combined with the smaller effective sample sizes within each subgroup. Thus, the hierarchical formulation appears clinically meaningful, but in its present form it does not yet provide a practical advantage over a direct ensemble model.

Several limitations of the study should be acknowledged. First, all models were implemented from scratch in NumPy. This ensured full transparency, which is a methodological strength, but it also limited computational efficiency and prevented the use of some advanced capabilities available in optimized libraries. Second, the feature space was restricted to routine cryoglobulin-related laboratory data. Important complementary information, including autoantibody profiles, complement fractions, viral load indicators, and structured clinical symptoms, was not available for modelling. Third, the dataset originated from a single institution, which inevitably constrains external generalizability. The clinical composition of the cohort, including local epidemiological and environmental factors, may differ from that of other centres and populations.

These limitations naturally point to future directions. A next step would be to reproduce the analysis using optimized implementations of modern boosting frameworks and to test whether the observed trends remain stable under stronger learners. Another important direction is feature enrichment, particularly through integration of immunological, virological, and symptom-based variables. External and multicentre validation will also be essential before any real deployment can be considered. Finally, future work may benefit from hybrid strategies that combine multiclass ranking, targeted pairwise discrimination, and calibrated uncertainty estimation within a single clinically oriented framework.

Overall, the findings suggest that machine learning may support exploratory analysis of cryopathy syndromes, but not in the form of a direct or clinically ready diagnostic tool. The most promising direction appears to be a layered modelling strategy that combines multiclass ranking, targeted binary clarification, and probability-aware interpretation. In this sense, the study provides initial benchmark results and outlines a cautious application perspective for future clinically oriented research.

## Conclusions

In this paper, the problem of machine learning-based classification of cryopathy syndromes from routine laboratory data was addressed. The analysis was performed under conditions of severe class imbalance, limited feature space, and substantial overlap between diagnostic categories. The obtained results allowed us to evaluate the proposed approaches and to assess their potential relevance for future clinically oriented decision-support research, and formulate the following main conclusions:

1. A comprehensive machine learning framework for the classification of cryopathy syndromes from routine laboratory data was developed and systematically evaluated. The study compared tree-based ensembles, neural networks, voting schemes, class-balancing strategies, hierarchical and period-aware models, targeted binary classifiers, probability calibration, and cross-validation procedures. Thus, the main aim of the study was achieved, and all planned methodological tasks were addressed within a single coherent framework.
2. To the best of our knowledge, we did not identify prior studies that addressed multiclass machine-learning classification of cryopathy syndrome categories from routine cryoglobulin-related laboratory data. In this sense, the present study helps fill an evident gap in the literature and provides an initial quantitative benchmark for this clinically important but computationally underexplored problem. This result is scientifically valuable because it offers a reference point for further studies in the field.

3. The comparative analysis showed that tree-based ensemble methods provide the most balanced behaviour in the multiclass setting. The best overall results were obtained by the soft-voting combination of Random Forest and Gradient Boosted Trees. This finding is methodologically important because it identifies a transparent reference model for future studies on cryopathy-oriented diagnostic support.

4. The study demonstrated that class balancing, feature engineering, and hyperparameter tuning substantially influence model behaviour. In particular, balancing strategies improved recognition of underrepresented classes, whereas stronger augmentation favoured dominant categories and improved overall accuracy at the cost of minority-class sensitivity. This result is important for practice because it shows that model quality in this problem must be assessed not only by global accuracy, but also by class-sensitive criteria.

5. A task-oriented combination of modelling strategies for highly imbalanced multiclass clinical data was examined. It integrates calibrated multiclass screening, Top-3 diagnostic shortlisting, and targeted binary differential classification within a single analytical concept. Its main value lies in aligning the computational pipeline with the general logic of staged clinical reasoning.

6. The analysis of advanced diagnostic strategies provided several original findings. The hierarchical approach showed that broader pathogenetic categories are easier to distinguish than final diagnoses, which confirms the internal structure of the problem. The period-aware analysis further showed that the relationship between cryoglobulin-related laboratory features and diagnosis remained broadly stable across the pre-COVID, COVID, and wartime periods. This is an important new result, especially in view of the unique wartime cohort included in the study.

7. The targeted binary classifiers for clinically relevant diagnostic pairs represented the most promising direction for future clinically oriented support within the present study. Their performance was generally stronger than in the full multiclass setting, which suggests that machine learning may be more useful as focused assistance in selected differential scenarios than as a direct one-step multiclass diagnostic tool.

8. The study also showed the value of calibrated and uncertainty-aware prediction. Probability calibration improved the interpretability of model outputs for the most represented classes, while shortlist-based prediction showed that the correct diagnosis often appears among the highest-ranked candidates even when it is not assigned the first position. This makes the proposed approach more suitable for preliminary shortlist generation in future decision-support research, where the system is intended to assist the physician rather than replace expert judgement.

9. A further contribution of the study is the identification and comparative evaluation of clinically informed engineered features that outperformed raw laboratory variables in the importance analysis. Features derived from cryoglobulin level and its interactions with hemagglutinin-related indicators proved especially informative. This finding has both methodological and practical importance, because it shows that domain-informed representation of laboratory data can reveal diagnostically relevant structure more effectively than the direct use of raw measurements alone.

Further research should focus on external and multicentre validation, expansion of the feature space with clinical, serological, and imaging variables, and implementation of the proposed framework using optimized modern libraries and more advanced tabular architectures. Particular attention should also be paid to hybrid decision-support strategies that combine multiclass ranking, targeted binary clarification, and calibrated uncertainty estimates within a single clinically oriented system.

**Ethics approval and consent to participate**

This study was conducted as a retrospective analysis of de-identified laboratory and diagnostic data collected in routine clinical practice. Ethical approval was granted by the Ethics Committee of Danylo Halytsky Lviv National Medical University (Approval No. 8, 29 August 2025).

**Consent for publication**

Not applicable.

**Availability of data and materials**

The datasets used and analysed during the current study are not publicly available because they contain sensitive clinical information, but de-identified data may be available from the corresponding author on reasonable request and subject to institutional and ethical approval. The code used in this study is available from the corresponding author on reasonable request.

**Competing interests**

The authors declare that they have no competing interests.

**Funding**

The authors received no specific funding for this work.

**Authors' contributions**

Nataliya Shakhovska, conceived the study, developed the methodology, performed the analysis, and drafted the manuscript. Valentyna Chopyak contributed to data curation, clinical interpretation, and manuscript revision. Ivan Izonin contributed to methodological validation and improvement of the manuscript. Vira Haievska contributed to data curation, clinical interpretation, and manuscript revision. All authors read and approved the final manuscript.

**Acknowledgements**

The authors would like to thank the reviewers for their careful reading of the manuscript and for their constructive comments and suggestions, which helped improve the quality and clarity of the paper.